\documentclass{article} 
\usepackage{iclr2025_conference,times}

\usepackage{amsmath,amsfonts,bm}

\def\eqref#1{equation~\ref{#1}}

\def\1{\bm{1}}

\DeclareMathAlphabet{\mathsfit}{\encodingdefault}{\sfdefault}{m}{sl}
\SetMathAlphabet{\mathsfit}{bold}{\encodingdefault}{\sfdefault}{bx}{n}

\usepackage{hyperref}
\usepackage{url}
\usepackage{algorithm}
\usepackage{algpseudocode}
\usepackage{spverbatim}
\usepackage{algpseudocode}
\usepackage{amssymb}
\usepackage{inconsolata}
\usepackage{booktabs}
\usepackage{multirow}
\usepackage{float}
\usepackage{svg}
\usepackage{longtable}

\title{CoheMark: A Novel Sentence-Level Watermark for Enhanced Text Quality}

\author{
Junyan Zhang\textsuperscript{1}, 
Shuliang Liu\textsuperscript{1}, 
Aiwei Liu\textsuperscript{2}, 
Yubo Gao\textsuperscript{1}, 
Jungang Li\textsuperscript{1}, 
Xiaojie Gu\textsuperscript{1}, 
Xuming Hu\textsuperscript{1}\thanks{Corresponding Author} \\
\textsuperscript{1}The Hong Kong University of Science and Technology (Guangzhou)\\
\textsuperscript{2}Tsinghua University \\
\texttt{junyanzhang0317@gmail.com} \\
}

\iclrfinalcopy 
\begin{document}

\maketitle

\begin{abstract}
Watermarking technology is a method used to trace the usage of content generated by large language models. Sentence-level watermarking aids in preserving the semantic integrity within individual sentences while maintaining greater robustness. However, many existing sentence-level watermarking techniques depend on arbitrary segmentation or generation processes to embed watermarks, which can limit the availability of appropriate sentences. This limitation, in turn, compromises the quality of the generated response. To address the challenge of balancing high text quality with robust watermark detection, we propose CoheMark, an advanced sentence-level watermarking technique that exploits the cohesive relationships between sentences for better logical fluency. The core methodology of CoheMark involves selecting sentences through trained fuzzy c-means clustering and applying specific next sentence selection criteria. Experimental evaluations demonstrate that CoheMark achieves strong watermark strength while exerting minimal impact on text quality.
\end{abstract}

\section{Introduction}

In recent years, the rapid advancement of large language models (LLMs) has revolutionized natural language processing \citep{openai2023gpt, yang2024qwen2, touvron2023llama}.
This technological leap, while marking a significant milestone in artificial intelligence, has also brought about unprecedented challenges \citep{xu2024hallucination, chen2023beyond, mazeika2024harmbench}. A major concern is that large language models can be exploited to generate false information and automated spam \citep{mirsky2023threat}.

To address this growing concern, researchers have begun focusing on developing various technologies to monitor AI-generated text and its usage. One effective way to track the usage of generated text is through watermarking, which involves embedding imperceptible information into the text \citep{kirchenbauer2023watermark, kuditipudi2023robust, zhao2023provable, giboulot2024watermax}. This makes it easier to detect and track the text for potential misuse. Compared to token-level watermarking methods, sentence-level watermarking is advantageous for preserving the internal semantic fluency within individual sentences and provides greater robustness. However, current sentence-level algorithms embed watermarks through arbitrary division of red-green region \citep{hou2023semstamp, hou2024k} following token-level works \citep{kirchenbauer2023watermark, zhao2023provable, hoang2024less}, which restricts the appropriateness of sentences, potentially introducing anachronistic or inappropriate tokens or sentences. This can reduce the overall quality of the generated content.
 
Therefore, to improve the issue of poor text quality while maintaining the effectiveness of watermark detection, we first deeply think about the intrinsic attributes of high-quality text. 
One key attribute of high-quality text is its strong coherence and fluency, which are essential for readability and comprehension \citep{xhepa2016importance}. Given the importance of coherence, what exactly characterizes a coherent and fluent text? 
According to cohesion theory \citep{halliday2014cohesion}, coherent and fluent texts establish connections between sentences through explicit and implicit linguistic means, such as lexical repetition, pronoun reference, synonyms, and contextually associated words, creating semantic overlap. Each sentence also relies on the preceding and following sentences for background information and contextual support \citep{ferstl2001role, xhepa2016importance}. Additionally, thematic consistency further enhances the semantic similarity between sentences. All of the above information highlights the semantic similarity and connection and that exists between sentences in high-quality text.

Finally, inspired by the principles, we propose CoheMark, which is an advanced technique that exploits the \textbf{Cohe}sive relationships between sentences for sentence-level water\textbf{Mark}. The process begins by dividing the semantic space using fuzzy c-means clustering \citep{dunn1973fuzzy, bezdek1984fcm}, which is an effective method for clustering sentence-level text \citep{skabar2011clustering}. 
In contrast to k-SemStamp proposed in \citet{hou2024k}, soft clustering offers two advantages over hard clustering. Firstly, in text clustering, some sentences may relate to multiple themes simultaneously, and soft clustering can more effectively capture this complexity. Furthermore, since sentences can exist across multiple semantic spaces, this clustering approach allows us to measure the degree to which a specific sentence aligns with various topical dimensions. It helps define criteria for selecting the next sentence and guides the semantic determination of the subsequent sentence by evaluating the topical relevance of the preceding sentence across different subject areas, ensuring proper text quality. Specifically, in comparison to earlier sentence-level watermarking techniques such as SemStamp and k-SemStamp \citep{hou2023semstamp, hou2024k}, our approach selects the next sentence by taking into account the semantics of the preceding sentence, thereby avoiding potential topic disruption or semantic disarray that may result from randomly choosing the valid semantic space for the next sentence.

We summarize our main contributions as follows: 
(1) We proposed CoheMark, a novel sentence-level watermarking method that leverages on semantic links between sentences. (2) We conducted extensive experiments across six baseline algorithms, two base LLMs, and two widely used datasets. The results indicate that CoheMark maintains comparable detectability and robustness against paraphrasing attacks compared to existing methods. (3) Compared to previous work, we undertook a more comprehensive evaluation of text quality using both traditional methods and LLM-based methods. Our findings demonstrate that CoheMark has a minimal impact on text quality and also achieves high performance in detection accuracy.

\section{Related Works}
\label{gen_inst}

\subsection{Watermarking for existing texts}
Watermarking existing text involves embedding hidden information without significantly changing the text's readability or meaning. This can be achieved through word-level modifications \citep{yang2022tracing, munyer2023deeptextmark, topkara2006hiding} or by altering the syntactic structure \citep{atallah2001natural, topkara2006words}, advancing watermarking methods for large language models.


\subsection{Watermarking for LLM-generated texts}
Watermarking LLM outputs has rapidly advanced in the past two years, with methods categorized into whitebox and blackbox models. Whitebox methods, often part of the KGW family \citep{kirchenbauer2023watermark}, split the vocabulary into green and red lists, favoring tokens from the green list during generation \citep{chen2024watme, hoang2024less, chen2024watme}. Other whitebox approaches apply watermarking during sampling \citep{kuditipudi2023robust, christ2024undetectable, hou2023semstamp, hou2024k}. In contrast, blackbox models, lacking access to logits or the ability to alter decoding, use lexical substitution or guide the insertion of ``secret'' words for watermarking \citep{qiang2023natural, yang2023watermarking, chang2024postmark}.

\subsection{Watermarked text quality-strength trade-off}

Watermarked text quality often trades off with watermark strength, as stronger watermarks can degrade quality \citep{molenda2024waterjudge}. Classic KGW \citep{kirchenbauer2023watermark} improves detectability but reduces text quality by producing infrequent tokens, while Unigram boosts robustness at the cost of diversity \citep{zhao2023provable}. CoheMark avoids altering the logit distribution, using sampling to balance quality and watermark strength effectively.

\begin{figure*}[h]
    \centering 
    \includegraphics[width=\textwidth]{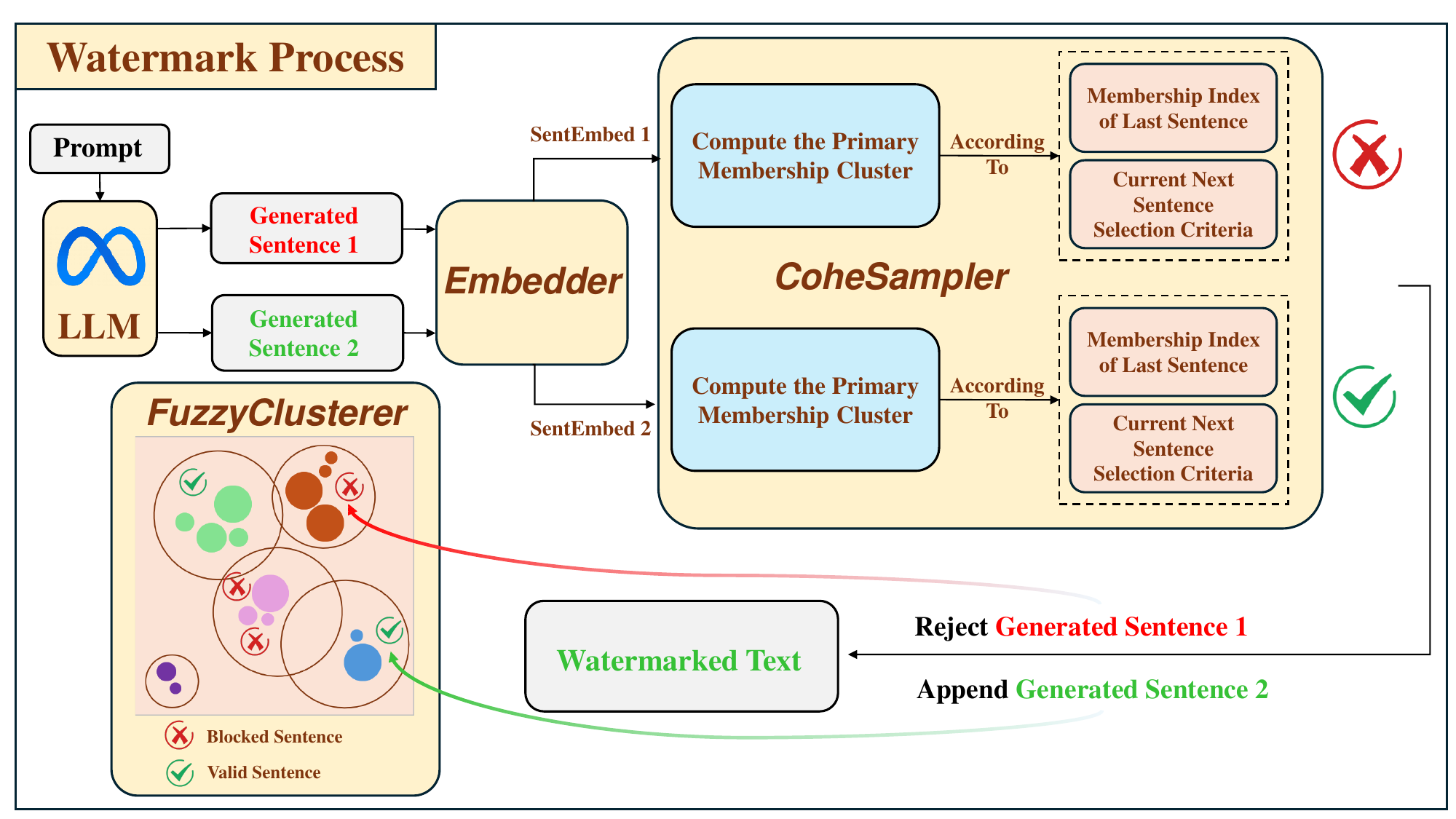} 
    \caption{The CoheMark watermarking procedure. Due to the inherent characteristic of rejection sampling within our watermarking framework, ``Generated Sentence 1'' exemplifies a textual output that fails to comply with the established watermark criteria and is consequently subjected to rejection. In contrast, ``Generated Sentence 2'' signifies a sample that has been accepted, adhering to the defined standards. Our watermarking methodology is fundamentally composed of three integral components: Embedder, FuzzyClusterer, and CoheSampler.}
    \label{fig:main} 
\end{figure*}

\section{CoheMark}
\label{headings}

We propose CoheMark, a technique that leverages the cohesive links between sentences for sentence-level watermarking. This method maintains a high detection rate while addressing the issue of low text quality. 
To implement CoheMark, we divide it into three modules---\textbf{Embedder}, \textbf{FuzzyClusterer} and \textbf{CoheSampler} to provide a clearer understanding of our watermark structure.
The rest of this section provides a detailed introduction to CoheMark.

\paragraph{Intuition}
Our intuition is that randomly dividing the red-green token or sentence spaces would significantly degrade the logical fluency of the text. Inspired by the characteristics of semantic coherence in text, we decided to design a watermarking scheme that leverages the cohesive relationships between sentences. This approach ensures that the division of valid and blocked regions considers the coherence of the whole text, thereby maintaining its quality and readability. Specifically, our method follows the process illustrated in Fig \ref{fig:main}.





%
\subsection{Inserting the watermark}

\paragraph{Embedder}
This is the embedding module. The Embedder's function is to project sentences into a high-dimensional latent space. 
This module is repeatedly used by FuzzyCluster and CoheSampler.

\begin{algorithm*}
\caption{CoheMark}
\begin{algorithmic}[!h]
\State \textbf{Input:} language model $M$, prompt $s^{prompt}$, Embedder $E$, trained fuzzy c-means clusters $C$, next sentence selection criteria $NSSC$.
    \For{$t = 1, 2, \dots, T$}
        \State 1. Compute the sentence embedding of the previously generated sentence $e^{t-1}$ = $E(s^{t-1})$ (for the first sentence, compute the embedding of the prompt).
        \State 2. Compute the Membership Index of $s^{t-1}$ using $C$. Then, according to the current version of $NSSC$, calculate the ``green semantic spaces'' $Green^{(t)}$ and the ``red semantic spaces'' $Red^{(t)}$ for the next sentence.
        \State 3. \textbf{repeat} Sample a new sentence $s^{t}$ from $M$, calculate the embedding $e^{t}$ = $E(s^{t})$ and compute the Primary Membership Cluster of $s^{t}$ to $C$. 
        \State \quad \textbf{until} 
        the Primary Membership Cluster is in the ``green semantic spaces'' $Green^{(t)}$.
        \State 4. Append the new sentence $s^{(t)}$ to generated context.
    \EndFor
\State \textbf{Output:} generated sequence $s^{(1)} \dots s^{(T)}$.
\end{algorithmic}
\end{algorithm*}

\paragraph{FuzzyClusterer}
This is the fuzzy c-means clustering module. This module performs fuzzy clustering on sentences from a specific domain, fuzzily partitioning the semantic space of different sentences to assist in the generation of watermarked text.
It allows us to effectively determine the semantic space from which we wish to sample the next sentence, based on the degree of membership in different semantic spaces exhibited by the previous sentence. In simpler terms, fuzzy clustering helps us choose the right topic or context for the next sentence by looking at how much the preceding sentence belongs to various topics.



Fuzzy c-means clustering is a soft clustering algorithm based on membership, which allows data points to belong to multiple clusters to varying degrees. Unlike traditional hard clustering methods (such as K-means and DBSCAN), it assigns a membership vector to each data point, which indicates the degree to which the data point belongs to each cluster. While extensively used in image segmentation and medical image analysis \citep{peng2024advanced, mohammed2020digital, chuang2006fuzzy}, fuzzy c-means clustering is also recognized as an effective technique for clustering sentence-level text \citep{skabar2011clustering}.

Specifically, in this module, we first define the specific domain for watermarking. Then, we sample $N$ sentences from this particular domain (e.g., news, Q\&A, finance), use Embedder to obtain sentence embeddings, and cluster them. 
The process begins with the initialization of the membership matrix and the cluster centers.
The membership matrix is updated using the following formula:
\begin{equation}
u_{ij} = \frac{1}{\sum_{k=1}^{C} \left( \frac{\| x_i - c_j \|}{\| x_i - c_k \|} \right)^{\frac{2}{m-1}} }
\end{equation}
Here, \( u_{ij} \) represents the degree to which data point \( x_i \) belongs to the \( j \)-th cluster, and \( m \) is the fuzziness parameter that controls the level of cluster fuzziness.
Next, the cluster centers are updated using the following formula:
\begin{equation}
c_j = \frac{\sum_{i=1}^{N} u_{ij}^m x_i}{\sum_{i=1}^{N} u_{ij}^m}
\end{equation}
In this equation, \( c_j \) represents the center of the \( j \)-th cluster. Iteration stops when changes in the membership matrix fall below a predefined threshold:
\begin{equation}
\sum_{i=1}^{n} \sum_{j=1}^{N} | u_{ij}^{(t+1)} - u_{ij}^{(t)} | \leq \epsilon
\end{equation}
where \( n \) is the number of data points, \( N \) is the number of clusters, and \( \epsilon \) is the threshold for terminating the iteration.
The FuzzyClusterer module is subsequently used by the CoheSampler.

%
%

\paragraph{CoheSampler}
This is the cohesive relation based sentence sampling module. This module is responsible for generating the main watermarked text.

Before explaining CoheSampler, we first introduce a concept called the Membership Index. After embedding a sentence using the Embedder $E$, we calculate the membership degree of this sentence to each cluster of the trained fuzzy c-means clusters $C$ obtained from FuzzyClusterer. The Membership Index represents the ranking of these membership degrees from highest to lowest. Formally, this can be expressed as:
\begin{equation}
    \text{Membership Index}(s^t) = \text{CalMem}(e^t, C)
    \label{eq:membership_index}
\end{equation}
where \( e^t \) is the embedding of the sentence \( s^t \) and \( C \) is the set of trained clusters.

For example, if there are 4 clusters, the Membership Index for a given sentence might be \{2, 3, 0, 1\}. This means that the sentence has the highest membership degree to the cluster indexed by 2, the second highest to the cluster indexed by 3, and so on. We define the cluster with the highest membership degree for a sentence as its Primary Membership Cluster.

After introducing the concepts of Membership Index and Primary Membership Cluster, we now describe the specific process of the watermark generation module, CoheSampler. In the generation of the $t$-th sentence, given the historical text $s^{(prompt)} \dots s^{(t-1)}$, where each $s$ represents a single sentence, we first compute the sentence embedding \( e^{t-1} \) of the previously generated sentence $s^{(t-1)}$ using the Embedder $E$. Then we compute the Membership Index of \( s^{t-1} \) using the trained fuzzy c-means clusters \( C \). Then, based on the current version of the next sentence selection criteria \( NSSC \), determine the ``green semantic spaces'' \( Green^{(t)} \) and the ``red semantic spaces'' \( Red^{(t)} \) for $s^t$. To generate the next sentence $s^t$, we continuously sample a new sentence from the language model $M$, embed the sentence using the same Embedder $E$ and calculate its Primary Membership Cluster until it is in the ``green semantic spaces'' $Green^{(t)}$.

In the preceding explanation, we referred to the statement ``current'' version of \( NSSC \). We provide further explaination here.
To effectively differentiate between watermarked and non-watermarked text, maintain a high detection rate, and simultaneously ensure high-quality text generation, we have decided to develop two distinct versions of the \( NSSC \).
Two versions are called $NSSC_{v1}$ and $NSSC_{v2}$, and we also define a Switching Rule between $NSSC_{1}$ and $NSSC_{v2}$. $NSSC_{v1}$ is primarily used, interspersed with the $NSSC_{v2}$. 
$NSSC_{v1}$ is mainly used to maintain the overall coherence of the semantics, while $NSSC_{v2}$ is used to appropriately pivot and update the semantics of the entire text, thereby enriching the content. 
For \( NSSC_{v1} \), the selection prioritizes alignment with the thematic context of the preceding sentence by evaluating the membership degrees of both the preceding and following sentences to various semantic spaces. Conversely, \( NSSC_{v2} \) employs a strategy that evaluates the degree of topical membership of the previous sentence across multiple themes, opting for modifications that introduce lesser-represented topics within the paragraph. The specific choices of $NSSC_{v1}$, $NSSC_{v2}$, and Switching Rule are detailed in the experimental setup.

\subsection{Detecting the watermark}
During the detection process, given a piece of text, our goal is to determine whether the text contains a watermark. Similar to the watermarking process, we use the same Embedder $E$ and the same trained fuzzy c-means clusters $C$ and apply the watermark rules, then calculate the watermark ratio $r$ to check what proportion of the sentences have been watermarked:
\begin{equation}
    r = \frac{S_V}{S_T}
    \label{eq:ratio}
\end{equation}
$S_V$ represents the number of sentences in the text that in the valid semantic spaces, and $S_T$ represents the total number of sentences in the text.
If the value of $r$ exceeds a certain threshold, it is likely that the text has been watermarked.

\section{Experiments}
\label{others}

In this section, we will introduce the experimental setup. Our Embedder is built upon Sentence-BERT \citep{reimers2019sentence}. In our experiments, we set the number of fuzzy c-means clusters to 8, as preliminary experiments indicated that this parameter setting achieves a good balance between watermark detection, text quality, and robustness.


\begin{table*}[!ht]
    \centering
    \caption{Comparison of CoheMark and baselines under detection performance and traditional text quality evaluation. All BertScores were multiplied by 100. The best and second-best performances are respectively highlighted in \textbf{bold} and \underline{underline}.}
    \label{main_table}
    \resizebox{1\textwidth}{!}{
    \begin{tabular}{lc|cccccc}
        \toprule
         &  & \multicolumn{3}{c}{OpenGen} & \multicolumn{3}{c}{LFQA} \\
        \cmidrule(lr){3-5} \cmidrule(lr){6-8}
        Model & Algorithm  & TPR@1\%  & PPL & BertScore  & TPR@1\% & PPL  & BertScore   \\
        \midrule
        \multirow{7}{*}{OPT-2.7B} & KGW         & 91.3  & 18.22 & 85.76  & 90.7  & 15.67 & 85.47  \\
         & KGW-2                                & 92.7  & 18.25 &  85.51  & 89.3 & 17.35 & 85.24  \\
         & KGW-4                                & 92.0  & 18.26 & 85.72   & 90.7 & 18.54 & 85.06    \\
         & EXP                                 & \textbf{100.0}  & \underline{17.84} &  85.50 & \textbf{100.0} & 29.76 & 84.70  \\
         & Unbiased                             & 92.7 & 18.31 &  85.56  & 90.0 & 17.06 & 85.27 \\
         & SemStamp                            & 98.3 & \textbf{17.49} & \textbf{86.02}   & 96.0 & \textbf{12.09} & \textbf{86.93} \\
         & \textbf{CoheMark}                    & \underline{99.3}  & 18.11 & \underline{85.98}  & \underline{97.3}  & \underline{14.48} & \underline{86.31}    \\
        \midrule
        \multirow{7}{*}{Llama-3-8B} & KGW         & 82.7 & \underline{9.40} & \textbf{86.13} & 97.3  & 5.83 & \underline{86.89}  \\
        & KGW-2                                 & 90.7 & 9.42 &  85.95 & \underline{99.3} & 5.76 & 86.81  \\
        & KGW-4                                 & 94.7  & 9.41 & \underline{86.12} & 98.7  & \underline{5.72} &  86.39    \\
         & EXP                                     & \textbf{100.0} & 9.52 &  85.63 & \textbf{100.0} & 18.41 & 85.66 \\
        & Unbiased                                 & 64.0 & \textbf{9.39} &  86.09  & 82.0 & \textbf{5.30} & \textbf{87.09}  \\
        & SemStamp                            & - & - &  -  & - & - & - \\
        & \textbf{CoheMark}                       & \underline{99.3}  & 9.45 & 86.04  & \textbf{100.0}  & 6.67 & 86.10   \\
        \bottomrule
    \end{tabular}
    }
\end{table*}

\paragraph{Next Sentence Selection Criteria:}
For $NSSC_{v1}$, we accept the sampling when the Primary Membership Cluster of the next sentence is in the first or third cluster in the Membership Index of the previous sentence. That is, the clusters indexed by one and three in the previous sentence's Membership Index are considered the ``green semantic spaces'' for the next sentence, while the rest are considered ``red semantic spaces''. For $NSSC_{v2}$, the second, fourth, fifth, and sixth clusters in the previous sentence's Membership Index are considered the ``green semantic spaces'' for the next sentence, while the rest are considered ``red semantic spaces''.


The Switching Rule is as follows: Initially, we select $NSSC_{v1}$. When there have been 5 cumulative instances where the Primary Membership Cluster of the previous sentence matches that of the next sentence, the next sample is chosen using $NSSC_{v2}$. After that, we switch back to $NSSC_{v1}$ until there is another instance of 5 cumulative matches between the Primary Membership Cluster of the previous and next sentences. We leave the optimization of the $NSSC$ and of the Switching Rule for future work.

\paragraph{Baselines:}
We compare CoheMark against four families of baseline algorithms, in total six baseline algorithms. 
(1) KGW (including KGW, KGW-2, and KGW-4) \citep{kirchenbauer2023watermark}. (2) EXP \citep{aaronson2023watermarking}. (3) Unbiased Watermark \citep{hu2023unbiased}. (4) SemStamp \citep{hou2023semstamp}. SemStamp encountered infinite loops for certain prompts using Llama-3-8B model, leading to excessive delays, so we excluded its results from the comparison. At present, our study has not undertaken a comparative analysis with k-SemStamp. This omission is due to the lack of k-SemStamp clustering outcomes to particular datasets. We propose to conduct a comparative evaluation involving k-SemStamp \citep{hou2024k} on the C4 \citep{raffel2020exploring} and BookSum \citep{kryscinski2021booksum} datasets, which the authors provide, as part of our future research endeavors.
For full details on all watermark methods settings, please refer to \S \ref{appendix: Details on baselines and hyperparameters}.


\paragraph{Metric for measuring detection performance:}
Following the work of previous researchers \citep{kirchenbauer2023watermark, hou2023semstamp, chang2024postmark}, we use binary classification metrics: true positive rate at false positive rates of 1\% (TPR@1\%). It means the percentage of correctly detected watermark text when 1\% of non-watermarked text are misclassified as watermarked text. 

\paragraph{Generative models:} 
In the experiment, we used the generative model OPT-2.7B \citep{zhang2022opt} and Llama-3-8B \citep{llama3modelcard}. Details on generative models can be found in \S \ref{appendix: Details on generative models}.



\paragraph{Datasets:} In our experiment, we utilized two specific datasets. (1) OpenGen, a dataset tailored for open-ended text generation \citep{krishna2024paraphrasing}. (2) LFQA dataset, another dataset collected by \citet{krishna2024paraphrasing}, which focuses on long-form question answering and includes a wide range of subject areas.


\subsection{Results of detectability}

In Table \ref{main_table}, CoheMark demonstrates highly competitive scores in terms of detection performance using rich unwatermarked text. It achieves over 97\% in TPR@1\% in both datasets across two different models. Compared to baseline methods, CoheMark and EXP exhibit comparable performance and outperform methods such as KGW, KGW-2, KGW-4, Unbiased watermark, and SemStamp. 

\subsection{Results of text quality}

Existing papers on watermarking often lack extensive and comprehensive quality assessment \citep{hoang2024less, kirchenbauer2023watermark, lu2024entropy}. However, we have taken a more comprehensive approach by including both traditional methods for evaluation and introducing large language models to thoroughly assess the quality of the text. Here, we aim to address two key questions:

\noindent\emph{{\textbf{Qusetion 1:}} How does CoheMark compare to other baselines in terms of text quality?}

\noindent\emph{{\textbf{Qusetion 2:}} Do traditional text quality evaluation tests really provide a objective and comprehensive assessment of text quality?}

\begin{table*}[!h]
    \centering
    \caption{Soft win rates calculated based on pairwise comparison evaluations, with GPT-4o serving as the judge. All numbers represent the soft win rate of CoheMark compared to the baseline models, where a value above 50 signifies CoheMark's superiority in the pairwise evaluation. ``Prefix Rel.'' refers to Relevance to the Prefix.}
    \label{tab:soft_win_rate}
    \resizebox{1\textwidth}{!}{
    \begin{tabular}{lc|ccccc}
        \toprule
         &  &  & &  CoheMark & &  \\
        \cmidrule(lr){3-7}
        Model & Algorithm & Overall & Prefix Rel. & Coherence & Interestingness &  Integrity \\
        \hline
        \multirow{4}{*}{OPT-2.7B}
        & KGW\textsubscript{swr}  & \textbf{79} & \textbf{73} & \textbf{77} & \textbf{83} & \textbf{80} \\
        & EXP\textsubscript{swr}  & \textbf{85} & \textbf{86} & \textbf{85} & \textbf{86} & \textbf{88} \\
        & Unbiased\textsubscript{swr} & \textbf{73} & \textbf{73} & \textbf{75} & \textbf{67} & \textbf{76} \\
        & SemStamp\textsubscript{swr} & \textbf{61} & \textbf{64} & \textbf{59} & \textbf{64} & \textbf{59} \\
                \hline
        \multirow{4}{*}{Llama-3-8B}
        & KGW\textsubscript{swr}      & \textbf{75} & \textbf{75} & \textbf{75} & \textbf{77} & \textbf{75} \\
        & EXP\textsubscript{swr}     & \textbf{95} & \textbf{95} & \textbf{95} & \textbf{97} & \textbf{93} \\
        & Unbiased\textsubscript{swr} & \textbf{70} & \textbf{76} & \textbf{73} & 43 & \textbf{76} \\
        & SemStamp\textsubscript{swr} & - & - & - & - & - \\
        \bottomrule
    \end{tabular}
}
\end{table*}

\paragraph{Traditional evaluation}
Following the majority of prior works \citep{kirchenbauer2023watermark, chen2024watme, hou2023semstamp}, we use perplexity to measure the quality of the generated text. 
To further assess the contextual semantic similarity and fluency of the generated text, we used BertScore \citep{zhang2019bertscore}. Specifically, in our experiment, we split the sentences within the watermarked text into many pairs of preceding and following sentences, and calculated the BertScore for each pair. The average BertScore of these pairs was used to represent the BertScore of the generated watermark. For example, if a watermarked text contains $n$ sentences, it would generate $n-1$ pairs, and the BertScore would be calculated $n-1$ times.

\paragraph{LLM-based evaluation}
In addition to traditional methods for evaluating text quality, approaches based on large language models have been proven to be effective or even more comprehensive than most existing automatic metrics \citep{chen2023exploring, zheng2023judging}. Thus, we utilize large language models to design the following pairwise preference evaluation.

\paragraph{Pairwise preference evaluation setup}
We use LLM as the judgement to perform paired comparison tasks. We have chosen GPT-4o as our evaluation criterion. Since KGW, KGW-2, and KGW-4 all belong to the KGW family, we selected watermarked text generated from KGW, EXP, Unbiased watermark, and SemStamp for our comparisons. Using the same 100 OpenGen prefixes, we generate watermark pairs with our method as well as with baseline methods. The model then assesses each response, gives its reasoning, and chooses its preferred response. We refer to \citet{lv2024longwanjuan}, extend the prompt introduced by \citet{chang2024postmark} and add the dimension of integrity. Therefore, the model is instructed to consider the relevance, coherence, interestingness, and integrity of responses when making judgments. For each metric, we record three types of outcomes: we win, the baseline wins, or it's a tie. The complete prompt can be found in \S\ref{appendix: pairwise}.
Subsequently, we calculated the \textbf{soft win rate} of CoheMark compared to other baseline methods in Table \ref{tab:soft_win_rate}. The soft win rate includes both the number of wins and the number of ties for CoheMark.

\paragraph{\emph{{\textbf{Qusetion 1:}}}}
Table \ref{main_table} compares the perplexity and BertScore metrics of texts generated using different baseline methods and CoheMark. It can be seen that when using the OPT model, CoheMark demonstrates competitive performance in both PPL and BertScore. However, when using Llama as the model, CoheMark outperforms EXP but is weaker than the KGW family methods and Unbiased watermark. We conducted a manual review of samples of generated watermark texts and found that the texts produced by CoheMark were superior to those of the baselines. This has led us to question whether traditional text quality evaluation strategies are truly fair, and has prompted us to continue using LLMs for a more comprehensive assessment of text quality.

Results from Table \ref{tab:soft_win_rate} show that CoheMark performs exceptionally well in the pairwise comparison across all baseline models. 
Regardless of whether the comparison is made against the OPT-2.7B or Llama-3-8B models, the text generated by CoheMark significantly outperforms that of the baseline models in terms of different metrics of text quality. Only in the case of the Llama models, CoheMark scores lower on the Interestingness metric compared to the Unbiased watermark.
The evaluation results of text quality based on state-of-the-art GPT-4o have conflicted with traditional assessment methods, which has prompted us to reflect. This will be discussed in detail in the following paragraph.

\paragraph{\emph{{\textbf{Qusetion 2:}}}}

Based on the evaluation by state-of-the-art LLM GPT-4o, we conjecture that traditional methods for assessing the quality of watermark generation may be relatively outdated and not comprehensive enough. For example, a generated text may perform well in terms of perplexity and BertScore, but it might lack practical meaning or logical coherence, or be filled with meaningless repetitions. This is because these two automatic evaluation strategies cannot fully capture the deep semantic and logical structure of the text. Meaningless repetitions can also lead to high perplexity and BertScore. We provide a specific example generated by KGW and EXP using Llama in \S\ref{appendix: repetition examples}. Therefore, we primarily rely on the LLM-based evaluation results, which indicate that CoheMark significantly outperforms other baselines in terms of quality of generated text.



%
\paragraph{Why baseline methods fail: case study}


To more clearly illustrate that CoheMark surpasses the baseline models in terms of generated text quality, we present a vivid example in Table \ref{tab: Generation Examples}. SemStamp was excluded from this analysis, as it yielded an empty string in response. 
We provided the following perspectives to explain why the text generated using the CoheMark method achieves superior results in LLM-based evaluations. 


First, compared to baseline models, CoheMark generates more coherent content, staying relevant to the prefix and maintaining internal coherence. For example, CoheMark's responses focus on Mount Elbert, discussing climbing and expeditions with a structured, chronological narrative. In contrast, KGW's responses jump between unrelated topics like camping, hiking routes, and summit facilities, leading to a disjointed output. Similarly, EXP veers into irrelevant information, mentioning unrelated locations like Zermatt Peak and the Subterranean River Sandstone. Unbiased Watermark also introduces broader geographical context in a similarly fragmented manner.


Second, compared to the baseline models, CoheMark is able to generate more accurate and reliable content. For example, when analyzing the response provided by CoheMark, although it also provides some fabricated historical details and incorrect elevation data, the overall content stays on topic. In contrast, the responses generated by KGW, EXP and Unbiased watermark introduce multiple factual errors and unrelated or fabricated details, containing numerous blatant inaccuracies, such as the existence of Mount Elbert State Park (which does not exist) and mentions of facilities like restrooms at the summit.

Third, compared to the baseline models, CoheMark is able to generate more engaging content. For example, when analyzing the response provided by CoheMark, it includes historical context and descriptive climbing-related anecdotes, which may engage a reader interested in Mount Elbert or climbing. In contrast, the content generated by KGW, EXP, and Unbiased watermark, with their large amount of fabricated and more generalized content, makes them very confusing.

\subsection{Results of robustness against attacks}

\begin{figure}[!h]
    \centering 
    \includegraphics[width=1\textwidth]{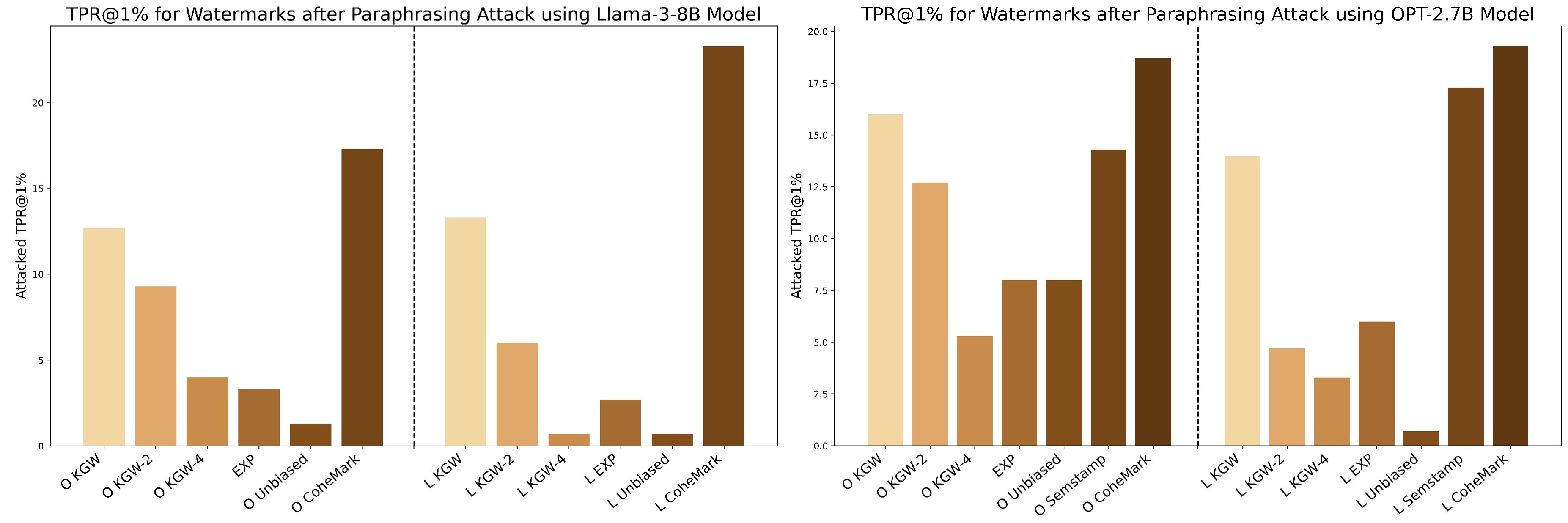} 
    \caption{Watermark robustness after sentence-level paraphrase attack. The symbols ``O'' and ``L'' in the figure correspond to the OpenGen and LFQA datasets, respectively. } 
    \label{tab:paraphrase} 
\end{figure}

Given that malicious actors can modify watermarked text sequences to evade detection, watermarking methods must ensure that the watermark remains robust against such alterations. We focus on semantic-preserving forms of attacks. Following prior works by \citet{hou2023semstamp, chang2024postmark, kirchenbauer2023reliability}, we use GPT-3.5-TURBO as our paraphrasing tool. Our approach involves sentence-level paraphrasing, where we give the model the prefix and the watermarked text and give the prompt to paraphrase sentence by sentence, . For more details on this setup, see \S \ref{appendix: paraphrase attack}. Our experimental results on the robustness to paraphrasing attacks are presented in Figure \ref{tab:paraphrase}. As illustrated, CoheMark demonstrates a higher True Positive Rate following a paraphrasing attack, outperforming other baseline models.

\section{Conclusion}
In this work, we propose a new sentence-level watermarking method called CoheMark, which exploits the cohesive relationships between sentences for watermarking. Through Embedder, FuzzyClusterer, and CoheSampler, CoheMark selects sentences from the ``green semantic region'' based on trained fuzzy c-means clusters and specific next sentence selection criteria.
Experimental results show that, on two datasets and two generative models, CoheMark maintains a high detection rate and watermark strength while addressing the issue of low text quality. 

\section{Limitations}
Although CoheMark achieves high performance in detection accuracy and has a minimal impact on text quality, there are some limitations that can be improved. Statistics reported in \S \ref{appendix: run times} show that, using the OPT-2.7B model on the OpenGen dataset, CoheMark's generation speed is approximately seven times slower than token-level watermarks. However, CoheMark's generation speed is still about three times faster than SemStamp, another sentence-level watermark. We view the additional computational cost of CoheMark as a trade-off for higher text quality and watermark robustness, despite the sacrifice in generation speed. Future work will focus on accelerating the generation speed of CoheMark, potentially through methods such as decoding multiple sentences simultaneously. 

Additionally, under the current settings, there is a small probability of generation failure, though this is very rare. Generation failure indicates its inability to produce unwatermarked text based on the provided prefix. This primarily occurs because the model consistently fails to generate a single sentence or keeps generating empty strings, thereby exceeding the maximum trial limit. Alternatively, it may be due to the inability to meet our next sentence selection criteria during the repetitive generations. We report the specific statistics in \S\ref{appendix: failure generations}. However, although this situation exists, the probability of it occurring is very low. We leave the optimization of generation failures to future work.

\bibliography{iclr2025_conference}
\bibliographystyle{iclr2025_conference}

\appendix

\section{Details on the settings of all algorithms}
\label{appendix: Details on baselines and hyperparameters}

For CoheMark, we conduct sampling at a temperature of 0.9 with a repetition penalty of 1.05 across two models. Additionally, we introduce a maximum number of trials. If the generation attempts for a particular sentence exceed this limit, the watermark text generation for that prompt will be halted. Among the possible scenarios, there may be cases where it is consistently unable to generate sentences that conform to specific semantic divisions, continuously fails to produce a single sentence, or keeps generating empty strings. This measure is primarily to prevent the watermark generation process from taking an excessively long time. All the baselines except Semstamp align with the settings in the MarkLLM \citep{pan2024markllm}. The prefix length is respectively set to 1, 2 and 4 for KGW, KGW-2 and KGW-4. 
For SemStamp, due to the unavailability of the fine-tuned model used in the study \citep{hou2023semstamp}, we employed the original model. We conducted sampling at a temperature of 0.7, while applying a repetition penalty of 1.05 to discourage repetitive sequences. The LSH dimension \( d \) was configured to 3, with a valid region ratio \( \gamma \) set to 0.25 and a rejection margin \( m \) established at 0.02.



\section{Details on generative models}
\label{appendix: Details on generative models}
During the generation process, following prior works \citep{chang2024postmark}, for all watermarking methods, we do not set a minimum number of tokens but set the maximum token limit to 255 for the OpenGen dataset and 120 for the LFQA dataset. That is because for the unaligned models, if allowed to generate freely until the end-of-sequence (EOS) token, it may produce meaningless repetitions, sometimes exceeding several thousand tokens. For the unwatermarked text, we use a uniform version generated by the aligned model Mistral-7B-Instruct-v0.3 \citep{jiang2023mistral}, using the same prompt and ask the aligned model to mimic human-written text and generate rich response.

The generative model checkpoints are: Facebook-OPT-2.7B and Meta-Llama-3.1-8B.

\section{Prompt for the LLM-based pairwise evaluation setup}
In this section, we present the prompt for the LLM-based pairwise evaluation setup. We extend the prompt introduced by \citet{chang2024postmark} by adding the dimension of integrity.
\label{appendix: pairwise}

\begin{spverbatim}
Please act as an impartial judge and evaluate the quality of the text completions provided by two large language models to the prefix displayed below. Assess each response according to the criteria outlined. After scoring each criterion, provide a summary of your evaluation for each response, including examples that influenced your scoring. Additionally, ensure that the order in which the responses are presented does not affect your decision. Do not allow the length of the responses to influence your evaluation. Be as objective as possible.

Criteria:
1. Relevance to the prefix
2. Coherence
3. Interestingness
4. Integrity

Start with a brief statement about which response you think is better or it is a tie overall. Then, for each criterion, state which response is better, or if there is a tie, followed by a concise justification for that judgment. At the very end of your response, declare your verdict by choosing one of the choices below, strictly following the given format:
[[A]] if Assistant A is better overall,
[[B]] if Assistant B is better overall,
[[C]] for a tie.

[Prefix]
{}

[Response A]
{}

[Response B]
{}

\end{spverbatim}




\section{Repetition and meaningless text examples}
\label{appendix: repetition examples}

We present a set of generation examples that exhibit repetition and lack meaningful content in Table \ref{tab: repetition and meaningless text}.

\begin{table}[h]
    \centering
    \caption{Generation pieces from OpenGen dataset generated by Llama-3-8B with different watermarks.}
    \label{tab: repetition and meaningless text}
    \resizebox{\columnwidth}{!}{
    \begin{tabular}{p{0.15\columnwidth} p{0.85\columnwidth}}
        \toprule
        KGW & Famous People Who Were Released From Prison Famous People Who Died in England List of Famous Londoners List of Famous Liverpool List of Famous Chester List of Famous Wirral People Famous People Who Were Killed During a Riot The Best Songs About London The Best Songs With Boogie in the Title The Best Songs with London in the Title Famous People Born in London The Best Music Videos of All Time The Best Songs in the World Ever  \\
        \midrule
        EXP     & Best domain name registrar in Hamburg, Missouri
Improve website to get more customers in Fort Wayne, Indiana
Digital marketing Fitchburg, Massachusetts
Best article marketing agency in Columbus, Mississippi
Digital marketing Beckley, West Virginia
Top SEO in Worcester, Massachusetts
Phone Message?
Your call will be returned in 1-2 business days.
This is   \\
        \bottomrule
    \end{tabular}
    }
\end{table}

\section{Generation examples}
In this section, we demonstrate watermarked examples using three watermarking algorithms using the prefix from OpenGen dataset, as shown in the Table \ref{tab: Generation Examples}.

\section{Paraphrase attack}
\label{appendix: paraphrase attack}
In this section, we offer a detailed explanation of the paraphrasing attack utilized in all our experiments. We expand upon the prompt introduced by \citet{hou2023semstamp, chang2024postmark} and guide a more potent paraphrasing attack:
\begin{spverbatim}
Given the previous prefix and the paragraph after that prefix, paraphrase the paragraph sentence by sentence. Only output the paraphrased paragraph in your response. Please maintain the core semantic meaning of each sentence from the original text. Feel free to modify the wording and sentence structure and try to replace adjacent vocabulary as much as possible to introduce new ways of expression, but do NOT change the primary information conveyed by the sentences.
Previous prefix: {}
Paragraph to paraphrase: {}
\end{spverbatim}

\section{CoheMark run times}
\label{appendix: run times}

In this section, we compare the runtime of CoheMark with several other baselines using OPT-2.7B as the generative model. Due to difficulties in running SemStamp \citep{hou2023semstamp}, another sentence-level watermark, we estimate the average generation time of SemStamp on the OpenGen dataset based on the information provided in \citet{hou2023semstamp}. According to the paper, SemStamp is 20.9 times slower than non-watermarked generation when using an LSH dimension \( d = 3 \) and a margin \( m = 0.02 \). Therefore, we estimate that SemStamp's average generation time on the OpenGen dataset is 77.12 seconds, which is approximately three times that of our method. As shown in Table \ref{tab:generation speed}, despite lagging behind token-level watermarks, our method, as a sentence-level watermark, offers faster generation speed compared to SemStamp.

We provide the following perspective to understand why CoheMark is generated faster than SemStamp, and why its generation speed is slower compared to token-level watermarks.
SemStamp randomly divides the valid and block regions of the next generated sentence based on the previous sentence, and sets a margin requirement to greatly increase the time required to generate each sentence. However, CoheMark adopts specific cohesive sampling rules, reducing the probability of rejecting samples. So the watermark sampling time is greatly shortened. As sentence level watermarks, both CoheMark and SemStamp have the characteristic of rejecting sampling, which makes sentence level watermarks slower than token level watermarks.

\section{CoheMark failure generation statistics}
\label{appendix: failure generations}
We present the failure generation statistics in Table \ref{tab:generation failure}.

\begin{table*}[!h]
    \centering
    \caption{Generation examples from OpenGen dataset generated by OPT-2.7B with different watermarks.}
    \label{tab: Generation Examples}   
    \scalebox{0.75}{
    \begin{tabular}{p{0.10\textwidth} p{0.90\textwidth}}
        \toprule
        Prompt & Mount Elbert is the highest summit of the Rocky Mountains of North America and the highest point in the U.S. state of Colorado and the entire Mississippi River drainage basin. The ultra-prominent fourteener is the highest peak in the Sawatch Range and the second-highest summit in the contiguous United States after Mount Whitney. Mount Elbert is located in San Isabel National Forest, southwest (bearing 223°) of the City of Leadville in Lake County, Colorado. The mountain was named in honor of a Colorado statesman, Samuel Hitt Elbert, who was active in the formative period of the state and Governor of the Territory of Colorado from 1873 to 1874. Henry W. Stuckle of the Hayden Survey was the first to record an ascent of the peak, in 1874.  \\
        \midrule
        CoheMark     & The peak\'s name was changed to Mount Elbert on December 13, 1890 through a joint resolution of the United States Congress. In 1892 a route was established by John William Stebbins around the northwest face of the mountain from the north side, and he made a successful climb the following year. A large section of the ridge between the head of the southern crater and Mount Marcy was named Stebbins Ridge in 1894. The Mount Marcy section was renamed after James W. Stebbins, a pioneer rock climber who successfully ascended the ridge in 1896. In 1899 Frederick B. Shaw led an advanced party by a new route up the west side of the mountain called the Golden Road from the northeast face. They climbed the peak via the "Golden Road" in 12 hours and 6 minutes and used the only route that would be possible today. Shaw claimed the peak as his own, naming it Shaw Peak in 1902. The name Mount Elbert was officially used again by the U.S. Army Ordnance Department to indicate the highest peak in the lower Mississippi River basin. The name became official when it was adopted by the American Alpine Club at its annual meeting in Denver, Colorado in 1907. The Mount Elbert area has been mapped as one of the premier climbing areas near Leadville and as recently as 1997 made the list of Colorado\'s "Top Ten Climbing Areas" by Adventure Sports magazine.   \\
                \midrule
        KGW     & The official mountain guide book, The Handbook to Colorado Mountaineering describes Elbert as "the most difficult peak in Colorado".

Mount Elbert State Park maintains a small visitors' center, with most information concerning activities of the nearby Leadville ski resort also located here. While there is no camping on the peak itself, camping is allowed in nearby state park campsites, but a permit is required. The "highest campground in North America", a  site called Horsethief, has been awarded the title of "highest dispersed camping" by Outdoor Retailer in 1998, and is a favorite among backcountry hikers. There are some facilities at the summit including restrooms; a post office is located nearby. Several routes ascend Elbert's south slope and provide spectacular alpine views over Colorado Springs to Puyallup and the Swan Mountains to Washington State. The steep northeast slopes are known as the "Eagle Route", one of the most difficult alpine routes on the planet, having received only two ascents since 2000. The summit, and the east face near the peak, is one of the highest points in North America not covered by glaciers. The peak has long attracted rock climbers, particularly via ferrata climbers, thanks to the sharp and dry cliffs that  
  \\
                \midrule
        EXP & Contents

Mount Elbert is one and two thirds miles in height above the combined elevation of its spires, with six high spires. Climbing South from Zermatt Peak it loses elevation 680 feet in less than a mile, by which time it has gained 1,075 feet. It stands at an altitude of 9,074 feet (2886.6 m), and marks the opening of the precipitous gorges known as "The Subterranean River Sandstone, or "The Hands." The southwest flank is twenty percent lower than the rest of the mountain, and most of the vertical rock is of this angle, showing a depth of over one hundred feet undertowed. A climb takes one hour, although the lariat ought to be made longer for contrast, and can be done wholly on the 'inside'. [1]

The summit faces North, pouring to the South as an almost perpendicular wall, over rocks exposed on the American side to an 
 \\
                 \midrule
         Unbiased & By 1907, nearly 100\% of the original recorded height had been reduced and, at that time, the name was reinstated by the U. S. Board of Geodetic Survey.

As of October 2018, the International Union of Mountain and Glaziers renamed the mountain as Colorado's highest peak in honor of climber John Longmont. The name honors Colorado's second mountain range, which includes both the eastern and western ranges, both of which are within the Sawatch mountain range.

Characteristics
Mount Elbert lies within the Sawattah Valley, a narrow valley in the south-southwest corner of the Sawatch Mountain Range. Mount Elberts highest point, the summit, is located at sea level,  to the south and  to the southwest of this prominence. Elevation from the ridge on the south is approximately, and elevation from the ridge along the ridge to the eastern summit is approximately 1,800 ft (525 m). The average summit slope is approximately 9\% with vertical relief ranging from 6 to 40 ft (2 to 12 m). About 30\% of the mountain is in the United States Forest Service's Mount Elbert Wilderness Area.

Overview
The mountain was one of Colorado's original seven mountains when Colorado
 \\
        \bottomrule
    \end{tabular}}
\end{table*}

\begin{table}[htbp]
        \vspace{-2cm}
    \centering
    \caption{Time of generation with OPT-2.7B as the base model. All numbers are computed over 10 generations.}
    \label{tab:generation speed}
    \begin{tabular}{lc}
        \toprule
        Algorithm & Average Runtime(s) \\
        \midrule
        KGW         &  4.18  \\
        KGW-2       &  3.91   \\
        KGW-4       &  3.84   \\
        EXP         &  3.65   \\
        Unbiased    &  4.51   \\
        SemStamp    &  77.12  \\
        CoheMark    &  27.40   \\
        \bottomrule
    \end{tabular}

\end{table}

\begin{table}[htbp]
        \vspace{-8cm}
    \centering
    \caption{Failure generation rates. All numbers are computed over 200 generations. ``FR'' refers to failure rates.}
    \label{tab:generation failure}
    \begin{tabular}{lcc}
        \toprule
        \multirow{2}{*}{Model} & OpenGen & LFQA\\
          & FR (\%) & FR (\%) \\
        \midrule
        OPT-2.7B         & 1.5 & 0.5  \\
        Llama-3-8B       & 4.5 & 1.0   \\
        \bottomrule
    \end{tabular}
\end{table}

\end{document}